\documentclass{article}

\PassOptionsToPackage{numbers, compress}{natbib}

\usepackage[preprint]{neurips_2025}

\usepackage[T1]{fontenc}    
\usepackage{hyperref}       
\usepackage{url}            
\usepackage{booktabs}       
\usepackage{amsfonts}       
\usepackage{nicefrac}       
\usepackage{microtype}      
\usepackage{xcolor}         
\usepackage{bbding}
\usepackage{array}
\usepackage{makecell}
\usepackage{tabularx}
\usepackage{bm}
\usepackage{multirow}
\usepackage{graphicx}
\usepackage{amsmath}
\usepackage{xcolor, color}
\usepackage{colortbl}
\usepackage{amssymb}
\usepackage{enumitem}
\usepackage{pifont} 
\usepackage{cellspace} 
\definecolor{highlight}{RGB}{58,110,165} 
\usepackage{caption} 
\usepackage{textcomp}
\usepackage{newunicodechar}
\newunicodechar{’}{'}
\newunicodechar{—}{---}
\newunicodechar{–}{--}
\newunicodechar{‐}{-}
\title{Generalizable Multispectral Land Cover Classification via Frequency-Aware Mixture of Low-Rank Token Experts}

\author{%
  Xi Chen \\
  National University of Defense Technology\\
  \texttt{xi\_chen@nudt.edu.cn} \\
  \And
  Shen Yan \\
  National University of Defense Technology\\
  \texttt{yanshen12@nudt.edu.cn} \\
  \And
  Juelin Zhu \\
  National University of Defense Technology\\
  \texttt{zhujuelin@nudt.edu.cn} \\
  \And
  Chen Chen \\
  National University of Defense Technology\\
  \texttt{chenchen16@nudt.edu.cn} \\
  \And
  Yu Liu \\
  National University of Defense Technology\\
  \texttt{jasonyuliu@nudt.edu.cn} \\
  \And
  Maojun Zhang\thanks{Corresponding author} \\
  National University of Defense Technology\\
  \texttt{mjzhang@nudt.edu.cn} \\
}

\begin{document}

\maketitle

\renewcommand*{\thefootnote}{\arabic{footnote}}
\setcounter{footnote}{0}


\begin{abstract}
We introduce Land-MoE, a novel approach for multispectral land cover classification (MLCC).
Spectral shift, which emerges from disparities in sensors and geospatial conditions, poses a significant challenge in this domain.
Existing methods predominantly rely on domain adaptation and generalization strategies, often utilizing small-scale models that exhibit limited performance.
In contrast, Land-MoE addresses these issues by hierarchically inserting a Frequency-aware Mixture of Low-rank Token Experts, to fine-tune Vision Foundation Models (VFMs) in a parameter-efficient manner.
Specifically, Land-MoE comprises two key modules: the mixture of low-rank token experts (MoLTE) and frequency-aware filters (FAF). MoLTE leverages rank-differentiated tokens to generate diverse feature adjustments for individual instances within multispectral images. By dynamically combining learnable low-rank token experts of varying ranks, it enhances the robustness against spectral shifts. Meanwhile, FAF conducts frequency-domain modulation on the refined features. This process enables the model to effectively capture frequency band information that is strongly correlated with semantic essence, while simultaneously suppressing frequency noise irrelevant to the task.
Comprehensive experiments on MLCC tasks involving cross-sensor and cross-geospatial setups demonstrate that Land-MoE outperforms existing methods by a large margin. 
Additionally, the proposed approach has also achieved state-of-the-art performance in domain generalization semantic segmentation tasks of RGB remote sensing images.
\end{abstract}

\section{Introduction}
\label{sec:intro}
Land cover classification aims to identify the land cover category corresponding to each pixel within remote sensing images~\cite{10377381}. This technique is critical in various fields, including geological exploration~\cite{eldosouky2024geological, li2024deep, hegab2024mineral}, wetland monitoring~\cite{sun2024monitoring, lin2024fuzzy, pan2024spatial}, urban planning~\cite{ning2024multi, zhu2024unrestricted, anand2024potential}, and precision agriculture~\cite{lei2025agricultural, jurado2022remote}.
Multispectral images (MSIs) are the preferred modality for this task, as they provide a broader range of spectral channels compared to RGB images and offer higher spatial resolution than hyperspectral images. 

\begin{figure}[tb]
    \centering
	\includegraphics[ width=1.0\linewidth]{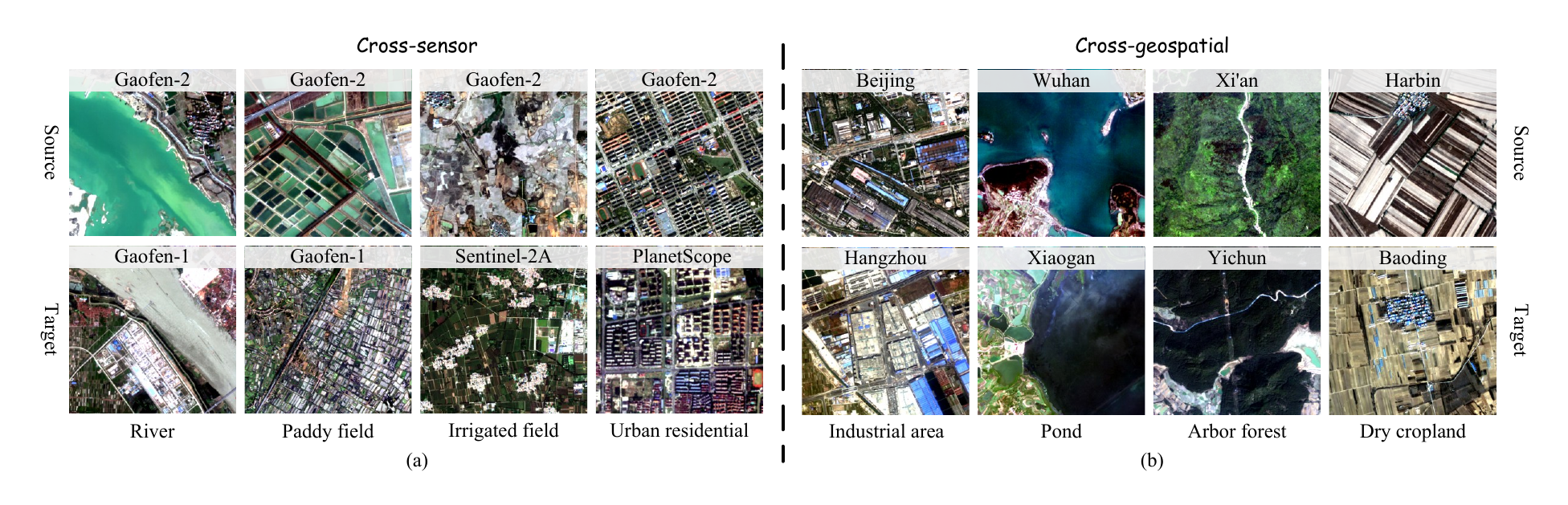}
	\caption{\textbf{Spectral shift in multispectral imagery.} Variations in sensor characteristics and geospatial conditions can lead to significant divergence in the spectral signatures of land cover features belonging to the same class.}
    \label{fig:introduction}
\end{figure}

Previous approaches~\cite{rad2024vision, noman2024rethinking, kodl2024arctic} for multispectral land cover classification (MLCC) often assume that source domain (SD) and target domain (TD) data are independently and identically distributed (IID).
However, real-world applications frequently encounter spectral shifts~\cite{10531019, ma2024transfer, ye2025towards} due to: 1) variations in sensor parameters across different multispectral sensors~\cite{xiao2024any} (i.e., cross-sensor), and 2) spatial heterogeneity in land cover type distributions and environmental lighting conditions across different geographic regions~\cite{song2024syntheworld, chen2024remote} (i.e., cross-geospatial), as illustrated in Figure~\ref{fig:introduction}. These shifts can substantially degrade model performance.

To improve generalization across diverse sensor and geospatial conditions, unsupervised domain adaptation (UDA)~\cite{broni2024unsupervised, 10497134, dionelis2024learning, wang2023unsupervised, hafner2022unsupervised, wang2021loveda} and domain generalization (DG)~\cite{10476506, ding2022domain, lang2024theoretical, deng2021scale, iizuka2023frequency} techniques have been proposed.
UDA methods utilize TD data to retrain models before testing to improve performance. While demonstrating competitive performance, this approach ties each model to a specific test scenario, necessitating retraining for every new scenario, thereby decreasing their efficiency in practical deployment~\cite{yue2023make, 10705912, guo2024start}. 
On the other hand, DG methods enhance model generalization through strategies like data augmentation~\cite{chattopadhyay2023pasta, li2021progressive, wang2021learning}, meta-learning~\cite{9157002, 10204921, wan2022meta}, and domain-invariant feature learning~\cite{xu2022dirl, huang2023idag}. However, these methods are often limited by the capacity of backbone networks to model the complex, non-linear relationships inherent in high-dimensional MSI. In real world, their generalization capabilities remain constrained.


Recent advances in dense image prediction tasks~\cite{chen2024practicaldg, su2024domain, li2024ada} have demonstrated the effectiveness of improving visual foundation models (VFMs)~\cite{radford2021learning, kirillov2023segment, fang2024eva, oquab2023dinov2} through  well-designed adapters~\cite{yi2024learning, wei2024stronger, bi2024learning}.
These approaches maintain VFM weights frozen while exclusively optimizing small inserted adapters, delivering excellent performance with minimal computational overhead~\cite{yu2024clipceil}.
In this study, we introduce this efficient fine-tuning paradigm to the MLCC for the first time. Our approach, called Land-MoE, innovates upon existing methods in adapter design: 1) We employ a Mixture-of-Experts (MoE) strategy to enhance generalization against spectral shifts. Unlike previous MoE implementations that utilize fully-connected or convolutional networks, we use learnable low-rank tokens with varied rank values as expert modules. By leveraging the collaborative interactions among multi-rank tokens, our method establishes pixel-level semantic associations and enhances the VFM's capacity for robust adaptation to various distribution shifts, while maintaining parameter efficiency through low-rank factorization. 2) We integrate frequency-aware filters that modulate the refined feature representations by preserving frequency components most pertinent to semantic content. This filter mechanism, being shared across layers for enhanced parameter efficiency, facilitates the model's ability to robustly capture semantic patterns across diverse scenes.

To validate our approach, we establish a new benchmark comprising diverse satellite imagery from globally distributed urban areas, evaluating the proposed method across multiple MLCC tasks under cross-sensor and cross-geospatial conditions. 
Experimental results demonstrate that Land-MoE significantly surpasses existing methods, both in accuracy and robustness.
Besides, the method achieves state-of-the-art results on RGB remote sensing imagery for land cover classification, underscoring its generalizability and potential for broader land cover prediction applications.

\paragraph{Contributions.}
\begin{itemize}[leftmargin=*]
\item First-time application of VFMs with efficient fine-tuning to MLCC tasks.
\item A new adapter with frequency-aware mixture of low-rank token experts to improve generalization.
\item State-of-the-art results across various sensor and geospatial conditions.
\end{itemize}

\section{Related Works}
\label{sec:relat_condensed}

\noindent \textbf{Generalizable Multispectral Land Cover Classification.}
Generalizable MLCC aims to enhance model generalization capabilities across domain distribution shifts~\cite{10646594}. Existing approaches primarily leverage UDA or DG. UDA methods often employ domain feature distribution alignment~\cite{10543066, 10966443,bai2024prompt} or self-training techniques~\cite{gao2025pseudo, hoyer2023mic, tong2020land, tong2023enabling}. While effective for a specific TD, UDA typically requires retraining for new, unseen scenarios. In contrast, DG methods train models exclusively on SD to generalize to unseen domains during testing~\cite{dong2023simmmdg}, utilizing strategies like data augmentation~\cite{chattopadhyay2023pasta, li2021progressive, wang2021learning}, domain-invariant feature learning~\cite{xu2022dirl, huang2023idag}, and meta-learning~\cite{9157002, 10204921, wan2022meta}. However, many DG methods are designed for RGB imagery and smaller backbone networks, which can limit their effectiveness in large-scale MLCC tasks. Our work seeks to bridge this gap by leveraging VFMs to achieve more practical and broadly generalizable MLCC without requiring domain-specific retraining.

\noindent \textbf{Parameter-efficient fine-tuning.}
State-of-the-art VFMs, often comprising billions of parameters~\cite{10834497}, present challenges for full fine-tuning due to prohibitive computational costs and potential performance degradation when task-specific data is limited compared to pre-training datasets~\cite{xin2024parameter}. Parameter-efficient fine-tuning (PEFT) addresses this by freezing most VFM parameters and optimizing only a minimal subset of task-specific parameters. This approach can achieve comparable or superior performance to full fine-tuning while significantly reducing resource consumption~\cite{ni2024pace}. In computer vision, mainstream PEFT approaches include adapter tuning, which integrates lightweight adaptable modules into transformer layers~\cite{kong2024moe, runwal2025peft, chen2024conv, chen2022adaptformer, jie2024convolutional, pan2022st}, and prompt tuning, which introduces learnable prompts into image embedding spaces~\cite{yang2024pedestrian, yao2023visual, li2023clip, jia2022visual, fang2024pros}. Differentiated from these, our approach introduces learnable low-rank token experts designed to dynamically adjust VFM features allocated to each expert.

\noindent \textbf{Mixture of Experts.}
The MoE paradigm enhances model capacity by dynamically combining multiple parallel expert subnetworks, where a routing network adaptively assigns expert weights based on input features~\cite{jacobs1991adaptive}. This mechanism has demonstrated notable performance advantages in natural language processing, computer vision, and recently in DG tasks~\cite{li2022sparse, lee2025domain}. However, current MoE frameworks often employ expert modules (e.g., fully-connected or convolutional networks) designed primarily for image-level classification. Such designs tend to overlook the fine-grained correlations between pixels, which are crucial for pixel-level classification tasks like MLCC. Our study addresses this limitation by proposing a novel method that employs learnable low-rank tokens with differentiated rank constraints as expert modules. By leveraging synergistic interactions among these multi-rank tokens, our method aims to improve the diversity of feature adjustments and establish robust pixel-level semantic relationships, thereby enhancing VFM domain generalization capabilities parameter-efficiently in complex multispectral remote sensing scenarios.
\section{Methodology}
\label{sec:method}

\subsection{Preliminary}
\begin{figure}[tb]
    \centering
	\includegraphics[ width=1.0\linewidth]{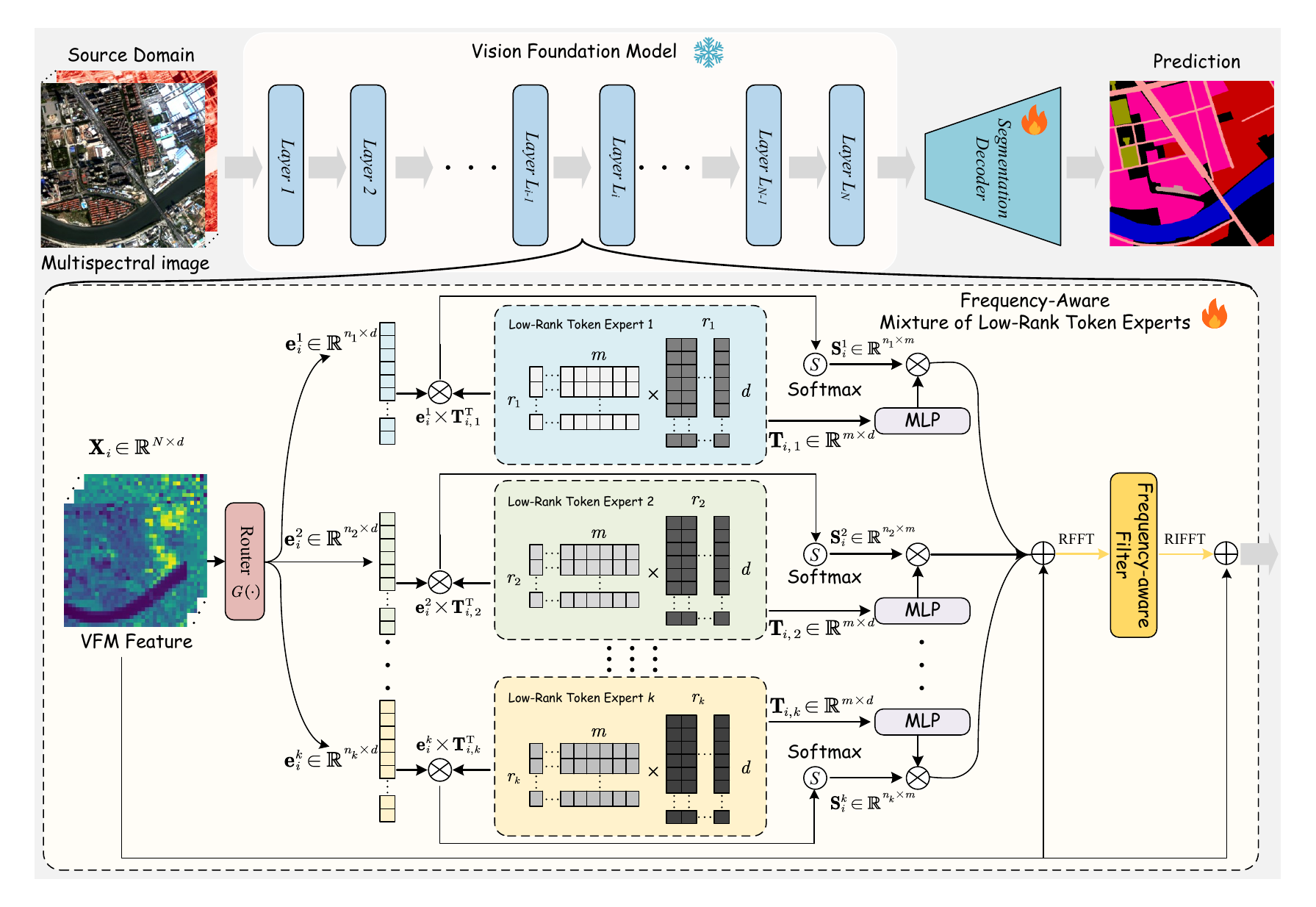}
	\caption{\textbf{Overview of Land-MoE}. 1. Land-MoE hierarchically inserts well-designed adapters into VFM backbone networks in a parameter-efficient manner to enhance their generalization for the cross-domain MLCC. 2. Land-MoE has two key modules, the Mixture of Low-rank Token Experts (MoLTE) and the Frequency-Aware Filters (FAF). 3. MoLTE enhances the adaptability of feature adjustments to spectral shifts by leveraging low-rank learnable token experts with varying ranks. 4. FAF performs frequency-domain modulation on the refined features output by the MoLTE module, perceiving frequency-domain features inherently correlated with semantic essence.}
    \label{fig:method}
\end{figure}

\paragraph{Problem formulation.}
We address large-scale MLCC under domain shift. We have a SD $\mathcal{D}_S = \{(\mathbf{X}^s_i, \mathbf{Y}^s_i)\}_{i=1}^{N_s}$ with $N_s$ annotated MSIs $\mathbf{X}^s_i \in \mathbb{R}^{H \times W \times C}$ and corresponding pixel-wise labels $\mathbf{Y}^s_i \in \{1, \dots, K\}^{H \times W}$, where $H, W$ are spatial dimensions, $C$ is spectral bands, and $K$ is class count. The TD $\mathcal{D}_T = \{\mathbf{X}^t_j\}_{j=1}^{N_T}$ contains $N_T$ unlabeled MSIs. Our objective is to train a model $f_{\theta}(\cdot)$ using $\mathcal{D}_S$ that exhibits strong generalization capabilities when applied to $\mathcal{D}_T$. 
The overall architecture of our proposed method designed to address this problem is illustrated in Figure~\ref{fig:method}.

\subsection{Architecture}
\label{subsec_Land-MoE}
\paragraph{Mixture of Low-rank Token Experts.}
To enable powerful VFMs to adapt to cross-scene MLCC tasks with enhanced generalization capabilities, Land-MoE first refines VFM features through a Mixture of Low-rank Token Experts (MoLTE). Specifically, the MoLTE consists of a routing network and a set of $N_e$ low-rank token experts, denoted by $\{E_j\}_{j=1}^{N_e}$, with varying ranks. Land-MoE employs a Top-$k$ noisy routing mechanism~\cite{chi2022representation} as its routing strategy, which dynamically injects input-dependent stochastic perturbations and sparsely activates domain experts. This design effectively mitigates spectral shift issues caused by geographical environmental variations and imaging condition differences in cross-scene MLCC.

For the feature sequence $\mathbf{X}_i\in \mathbb{R} ^{N\times d}$ output by the $i$-th layer of the VFM, where $N$ is the number of tokens and $d$ is the feature dimension, the routing score $G\left( \mathbf{X}_i \right) \in \mathbb{R}^{N \times N_e}$ is computed for each token $\mathbf{x}_{i,n} \in \mathbb{R}^d$ ($n=1, \dots, N$):
\begin{equation}
 \label{eq:1}
 G\left( \mathbf{x}_{i,n} \right) =\mathrm{Softmax} \left( \mathrm{Topk}\left( H\left( \mathbf{x}_{i,n} \right) ,k \right) \right) \in \mathbb{R}^{N_e}
\end{equation}
The $\mathrm{Topk}$ operation selects the top-$k$ scores for the current token and sets the remaining $N_e-k$ scores to $-\infty$. The function $H\left( \mathbf{x}_{i,n} \right) \in \mathbb{R} ^{N_e}$ dynamically determines the base routing scores allocated to different low-rank token experts for the input token $\mathbf{x}_{i,n}$, which can be formally expressed as:
\begin{equation}
 \label{eq:2}
 H\left( \mathbf{x}_{i,n} \right) =\mathbf{x}_{i,n}\mathbf{W}_{i}^{g}+\epsilon \odot \mathrm{Softplus}\left( \mathbf{x}_{i,n}\mathbf{W}_{i}^{noise} \right)  
\end{equation}
where $\epsilon \sim \mathcal{N} \left( 0,1 \right) ^{N_e}$ denotes a vector of noise sampled from the standard normal distribution, $\mathrm{Softplus}\left( z \right) =\log \left( 1+\exp \left( z \right) \right) $ is the activation function applied element-wise, and $\mathbf{W}_{i}^{noise}\in \mathbb{R} ^{d\times N_e}$ and $\mathbf{W}_{i}^{g}\in \mathbb{R} ^{d\times N_e}$ are learnable weight matrices in the routing network. $\odot$ denotes the Hadamard product.

After assigning each token $\mathbf{x}_{i,n}$ in $\mathbf{X}_i$ to its corresponding Top-1 low-rank token expert $E_{i, \text{top1}}\left( \cdot \right)$ based on the routing scores $G\left( \mathbf{x}_{i,n} \right)$, let $\mathbf{e}_{i,n} = \mathbf{x}_{i,n}$ be the patch feature assigned to expert $E_{i, \text{top1}}$. This feature is refined using the learnable low-rank tokens $\mathbf{T}_{i, \text{top1}}\in \mathbb{R} ^{m\times d}$ within the expert $E_{i, \text{top1}}$. The correlation $\mathbf{S}_{i,n} \in \mathbb{R}^{1 \times m}$ between the assigned patch feature $\mathbf{e}_{i,n}$ and each learnable low-rank token in $\mathbf{T}_{i, \text{top1}}$ is computed as:
\begin{equation}
 \label{eq:3}
 \mathbf{S}_{i,n}=\mathrm{Softmax} \left( \frac{\mathbf{e}_{i,n}\mathbf{T}_{i, \text{top1}}^{\mathrm{T}}}{\sqrt{d}} \right)  
\end{equation}
where $m$ denotes the number of learnable tokens per expert, and $d$ is their dimension. Subsequently, the learnable low-rank tokens $\mathbf{T}_{i, \text{top1}}\in \mathbb{R} ^{m\times d}$ are projected into the feature space of $\mathbf{e}_{i,n}$ via a multi-layer perceptron (MLP). These projected tokens are then weighted by the correlation map $\mathbf{S}_{i,n}$ to derive the adjustment term $\Delta \mathbf{e}_{i,n} \in \mathbb{R}^d$ for the assigned feature $\mathbf{e}_{i,n}$. This procedure is formulated as:
\begin{equation}
 \label{eq:4}
 \Delta \mathbf{e}_{i,n}=\mathbf{S}_{i,n} \left( \mathbf{T}_{i, \text{top1}}\mathbf{W}_T+\mathbf{b}_T \right)  
\end{equation}
where $\mathbf{W}_T\in \mathbb{R} ^{d\times d}$ and $\mathbf{b}_T\in \mathbb{R} ^{m\times d}$ denote the weights and biases of the MLP, respectively. The MLP uses layer-wise weight sharing across experts. The adjustment terms $\Delta \mathbf{e}_{i,n}$ for all tokens are then aggregated to form $\Delta \bar{\mathbf{X}}_i \in \mathbb{R}^{N \times d}$, where each row corresponds to the adjustment for the respective token.

\paragraph{Frequency-aware Filters.}
Building upon the refined features from the MoLTE, $\Delta \bar{\mathbf{X}}_i+\mathbf{X}_i$, we further establish explicit associations between class-semantic-related features and frequency-domain components through frequency-domain analysis and dynamic filtering. This achieves adaptive enhancement of semantically essential frequency features. Specifically, we first apply the real-valued fast Fourier transform (RFFT) to the MoLTE-refined features. Let $\mathbf{Z}_i = \Delta \bar{\mathbf{X}}_i+\mathbf{X}_i$. Assuming $\mathbf{Z}_i$ can be reshaped or processed as a spatial grid of size $h \times w$ for frequency analysis, the frequency-domain representation $\mathcal{F} \left( \mathbf{Z}_i \right) \in \mathbb{C}^{h \times (\lfloor w/2 \rfloor + 1) \times d}$ is formally described as:
\begin{equation}
 \label{eq:5}
 \mathcal{F} \left( \mathbf{Z}_i \right) =\mathrm{RFFT}\left( \mathbf{Z}_i \right)  
\end{equation}
where RFFT denotes the real-valued fast Fourier transform operation applied channel-wise for each spatial location. RFFT is used to preserve spectral amplitude information while avoiding conjugate symmetry redundancy, thereby reducing the number of learnable parameters in frequency-aware filters.

Subsequently, frequency filtering is performed via a learnable frequency-domain filter $\mathbf{W}_{filter}$ to amplify semantically relevant frequency components and suppress noise within the spatially refined features. This process is formulated as:
\begin{equation}
 \label{eq:6}
 \hat{\mathcal{F}}\left( \mathbf{Z}_i \right) =\mathcal{F} \left( \mathbf{Z}_i \right) \odot \mathbf{W}_{filter}  
\end{equation}
where $\mathbf{W}_{filter}\in \mathbb{R} ^{h\times \left( \lfloor \frac{w}{2} \rfloor +1 \right) \times d}$ denotes the learnable frequency-aware filter, $\odot $ represents the Hadamard product. The filter weights are shared across layers to reduce the number of learnable parameters. After frequency modulation, the filtered frequency-domain features $\hat{\mathcal{F}}\left( \mathbf{Z}_i \right)$ are transformed back to the spatial domain via the real-valued inverse fast Fourier transform (RIFFT). This process is formalized as:
\begin{equation}
 \label{eq:7}
 \Delta \mathbf{X}_i=\mathrm{RIFFT}\left( \hat{\mathcal{F}}\left( \mathbf{Z}_i \right) \right)  
\end{equation}
where $\mathrm{RIFFT}$ denotes the real-valued inverse fast Fourier transform operation, and $\Delta \mathbf{X}_i \in \mathbb{R}^{N \times d}$ represents the final feature adjustment term from this module, which is then typically added back to $\mathbf{X}_i$.

\subsection{Details of Land-MoE}
\label{subsec_Details_of_Land-MoE_condensed}
\paragraph{Layer-wise feature refinement.}
Land-MoE enhances VFM generalization by refining features layer-wise. For each of the $L_N$ VFM layers where Land-MoE is applied, its module processes the $i$-th layer's output feature sequence $\mathbf{X}_i \in \mathbb{R}^{N \times d}$ to produce an adjustment term $\Delta \mathbf{X}_i$. The input to the subsequent $(i+1)$-th layer $f_{i+1}$ is then the refined feature $\mathbf{X}_i + \Delta \mathbf{X}_i$. This iterative process is described as:
\begin{equation}
\label{eq:8_condensed}
\mathbf{X}_{i+1}=f_{i+1}\left( \mathbf{X}_i+\Delta \mathbf{X}_i \right), \quad i=1,2,\dots, L_N-1
\end{equation}

\paragraph{Learnable low-rank tokens experts.}
The MoLTE component at layer $i$ utilizes $N_{e,i}$ learnable token experts, denoted by $\{\mathbf{T}_{i,k} \in \mathbb{R}^{m \times d}\}_{k=1}^{N_{e,i}}$, where $m$ is the number of learnable tokens per expert. To enhance feature representation diversity and significantly reduce learnable parameters, each expert $\mathbf{T}_{i,k}$ is low-rank factorized:
\begin{equation}
\label{eq:9_condensed}
\mathbf{T}_{i,k}=\mathbf{A}_{i,k}\mathbf{B}_{i,k}
\end{equation}
where $\mathbf{A}_{i,k} \in \mathbb{R}^{m \times r_k}$ and $\mathbf{B}_{i,k} \in \mathbb{R}^{r_k \times d}$ are factor matrices, and $r_k$ is the rank satisfying $r_k \ll \min(m,d)$. The learnable parameters in MoLTE are primarily these low-rank matrices $\{\mathbf{A}_{i,k}, \mathbf{B}_{i,k}\}$.

\paragraph{Optimization objective.}
To adapt VFMs for MLCC, we use the Mask2Former loss $\mathcal{L}_{Mask2former}$ as our primary semantic learning objective. Additionally, to encourage a balanced selection of experts by the MoLTE routing network, we introduce an expert balancing loss $\mathcal{L}_{MoLTE}$:
\begin{equation}
\label{eq:10_condensed}
\mathcal{L}_{MoLTE}=\sum_{j=1}^{N_L} \sum_{k=1}^{N_{e,j}} \left( \frac{\mathrm{std}_{\mathbf{X}\in \mathcal{B}} \left( \sum_{\mathbf{x}_{n} \in \mathbf{X}} G_{j,k}\left( \mathbf{x}_n \right) \right)}{\mathrm{mean}_{\mathbf{X}\in \mathcal{B}} \left( \sum_{\mathbf{x}_{n} \in \mathbf{X}} G_{j,k}\left( \mathbf{x}_n \right) \right) } \right)^2
\end{equation}
where $\mathcal{B}$ is a batch of input feature sequences, $N_L$ is the number of Land-MoE layers, $N_{e,j}$ is the number of experts in MoLTE layer $j$, and $G_{j,k}(\mathbf{x}_n)$ is the routing score from token $\mathbf{x}_n$ to expert $k$ in layer $j$. This loss penalizes high variance in the total routing mass assigned to each expert across a batch. The final optimization objective is a weighted sum:
\begin{equation}
\label{eq:11_condensed}
\mathcal{L} =\mathcal{L} _{Mask2former}+\lambda \mathcal{L} _{MoLTE}
\end{equation}
where $\lambda \ge 0$ is a hyperparameter controlling the contribution of the expert balancing loss.

\section{Experiments}
\label{sec:exp}
Extensive experiments are conducted on the cross-sensor and cross-geospatial tasks to demonstrate the effectiveness of our proposed Land-MoE as described in Sec.~\ref{exp:compare_sota}. Additionally, ablation studies are conducted on the cross-sensor and cross-geospatial tasks in Sec.~\ref{exp:ablation_study}. More results, including further parameter analysis and Land-MoE's generalization performance on RGB remote sensing images, are provided in Appendix~\ref{sec:detail_param_analysis} and Appendix~\ref{sec:performance_RGB_generalization}, respectively.

\begin{figure}[htbp]
    \centering
	\includegraphics[ width=1.0\linewidth]{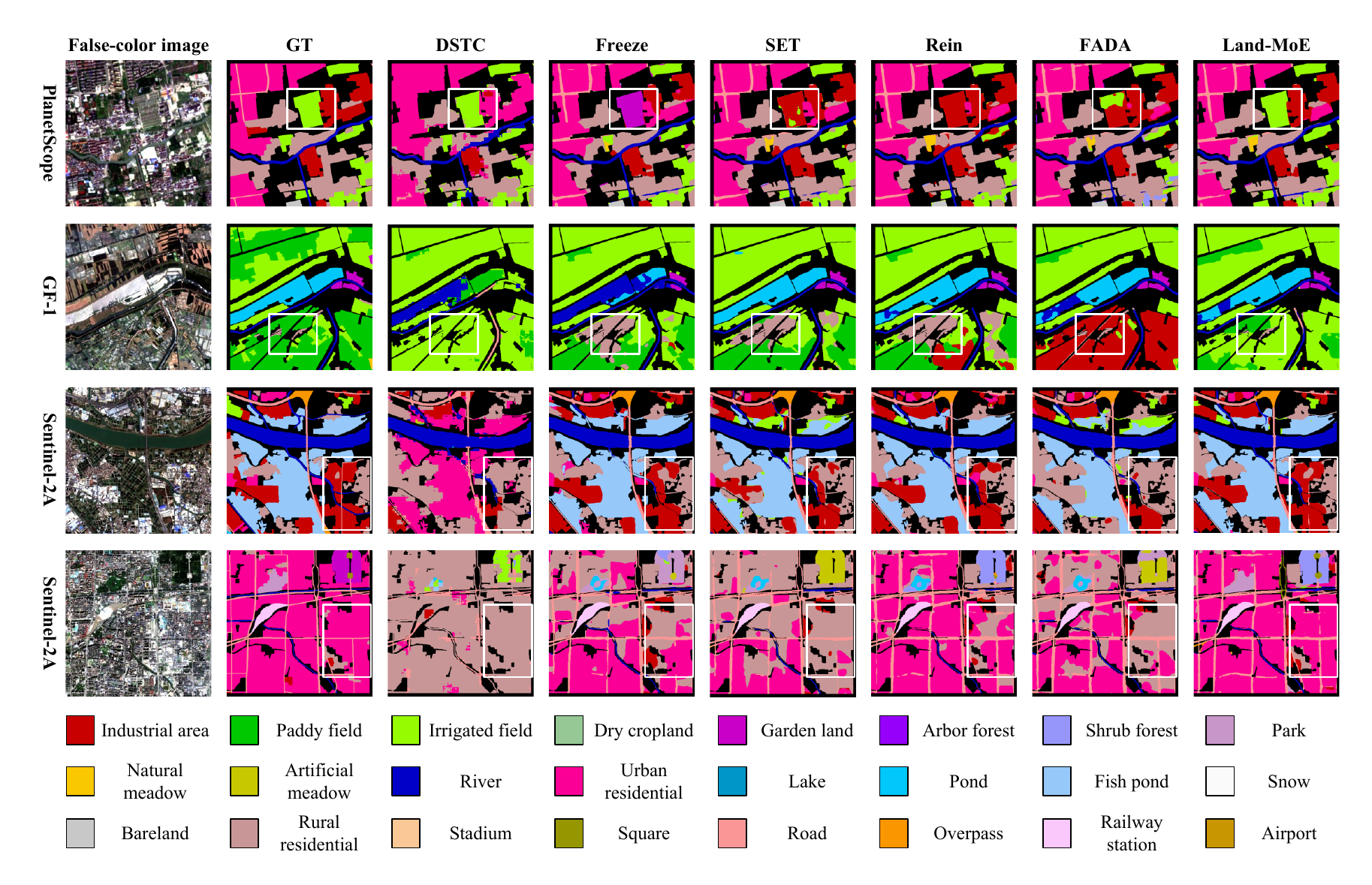}
	\caption{\textbf{Qualitative results for cross-sensor MLCC task.} Comparative visualization of land cover classification from the IID-based method DSTC~\cite{liu2024dual}, frozen DINOv2 + Mask2Former decoder, VFM-based DG semantic segmentation methods (SET~\cite{yi2024learning}, Rein~\cite{wei2024stronger}, FADA~\cite{bi2024learning}), and our proposed Land-MoE. Input MSIs and corresponding ground truth maps are also shown for reference. Land-MoE exhibits  superior accuracy in challenging cross-sensor scenarios. Please zoom in to the white box region to see more details.}
    \label{fig:Cross_sensor}
\end{figure}

\subsection{Evaluation Protocols}
\noindent \textbf{Datasets.}
Our experiments establish cross-sensor and cross-geospatial generalization tasks based on GF-2 MSIs from the Five-Billion-Pixels dataset~\cite{tong2023enabling}. For the cross-sensor task, these GF-2 MSIs serve as the SD, while MSIs from GF-1, PlanetScope, and Sentinel-2 form the TDs. For the cross-geospatial task, we partition the GF-2 MSIs within the Five-Billion-Pixels dataset into geographically disjoint SD and TD. Please refer to Appendix~\ref{sec:detail_dataset} for more details.

\noindent \textbf{Baselines.}
We compare the proposed method with the following baselines in two categories: 1) state-of-the-art MLCC method that assume IID, namely DSTC~\cite{liu2024dual}. 2) VFM-based semantic segmentation methods for domain generalization, including SET~\cite{yi2024learning}, Rein~\cite{wei2024stronger}, and FADA~\cite{bi2024learning}.
For fair evaluation, all compared VFM-based methods utilize the same VFM and decoder from ours and are retrained with optimal parameters from their original works.


\noindent \textbf{Implementation details.}
We employ DINOv2~\cite{oquab2023dinov2} as the default VFM backbone and the widely adopted Mask2Former~\cite{cheng2022masked} decoder for pixel-wise land cover classification. Notably, the proposed Land-MoE is compatible with other advanced VFMs. MSIs are preprocessed by cropping to $512 \times 512$. For fair comparison, data augmentation strategies are kept consistent with compared baselines (SET~\cite{yi2024learning}, Rein~\cite{wei2024stronger}, FADA~\cite{bi2024learning}). Training utilizes AdamW optimizer ($1\times 10^{-4}$ initial learning rate, batch size 8) for 20 epochs. All experiments were conducted on NVIDIA RTX 4090 GPUs.

\noindent \textbf{Metrics.}
The proposed method is quantitatively evaluated against other baselines using mean accuracy (mAcc), mean intersection over union (mIoU), and per-class accuracy metrics.


\subsection{Evaluations}
\label{exp:compare_sota}

\noindent \textbf{Evaluation on cross-sensor tasks.}
The rationale for this experiment stems from the inherent limitations of relying on a single satellite sensor, which is often constrained by cloud occlusion and prolonged revisit cycles.
As demonstrated in Table~\ref{tab:cross_sensor}, Land-MoE outperforms both the IID-based DSTC method~\cite{liu2024dual} and vision VFM-based DG semantic segmentation approaches, including SET~\cite{yi2024learning}, Rein~\cite{wei2024stronger}, and FADA~\cite{bi2024learning}. DSTC exhibits the lowest performance, attributable to its reliance on the IID assumption. Frozen VFM(DINOv2) yields a substantial improvement (+35.02\% mIoU over DSTC), underscoring its inherent generalization capabilities. State-of-the-art VFM-based DG semantic segmentation methods can further enhance segmentation accuracy. Notably, Land-MoE achieves superior results, exceeding SET, Rein, and FADA by 8.63\%, 5.30\%, and 7.79\% mIoU, respectively, validating its robustness in challenging cross-sensor scenarios.

\begin{table}[htbp]
\centering
\caption{\textbf{Cross-sensor Performance.} Results (mAcc, mIoU, per-class acc.\% for 24 classes) comparing Land-MoE and baselines. Land-MoE shows superior overall performance.}
\label{tab:cross_sensor}
\footnotesize
\renewcommand{\arraystretch}{1.1}

\setlength{\tabcolsep}{3pt} 
\begin{tabular}{@{} l c c *{10}{c} @{}} 
\toprule
\multicolumn{1}{c}{\textbf{Method}} &
\multicolumn{2}{c}{\textbf{Overall}} &
\multicolumn{10}{c}{\textbf{Per-Class Accuracy (\%)}} \\
\cmidrule(lr){2-3} \cmidrule(l){4-13}
& {mAcc} & {mIoU} & {C1} & {C2} & {C3} & {C4} & {C5} & {C6} & {C7} & {C8} & {C9} & {C10} \\
\midrule

DSTC\scriptsize{(ECCV'24)} & 40.38 & 29.34 & 40.49 & 34.47 & 86.13 & 0.00 & 0.00 & 94.94 & 0.00 & 0.77 & 31.28 & 27.46 \\
DINOv2\scriptsize{(Freeze)} & 69.10 & 53.37 & 64.13 & 55.96 & 91.15 & 0.00 & 27.06 & 93.98 & 50.23 & 55.17 & 80.97 & 42.50 \\
SET\scriptsize{(MM'24)} & 70.24 & 55.73 & 64.18 & 75.17 & 90.55 & 0.00 & 45.47 & 94.66 & 30.13 & 77.08 & 80.46 & 66.17 \\
Rein\scriptsize{(CVPR'24)} & 73.44 & 59.06 & 66.78 & 75.22 & 90.00 & 0.00 & 43.03 & 97.08 & 28.27 & 81.77 & 80.76 & 39.89 \\
FADA\scriptsize{(NeurIPS'24)} & 73.21 & 56.57 & 68.74 & 70.01 & 88.19 & 0.00 & 47.09 & 97.92 & 74.00 & 73.83 & 81.99 & 55.46 \\
\rowcolor{gray!10}
\textcolor{highlight}{\textbf{Land-MoE\scriptsize{(Ours)}}} & \textbf{77.95} & \textbf{64.36} & 73.27 & 72.67 & 92.31 & 0.00 & 51.43 & 97.65 & 30.44 & 84.82 & 88.42 & 64.82 \\ 
\bottomrule
\end{tabular}

\vspace{1pt} 

\setlength{\tabcolsep}{3.5pt} 
\begin{tabular}{@{\hspace{1pt}} *{14}{c} @{\hspace{1pt}}} 
\toprule
\multicolumn{14}{c}{\textbf{Per-Class Accuracy (\%) (Continued)}} \\
\cmidrule(l{5pt}r{5pt}){1-14} 

{C11} & {C12} & {C13} & {C14} & {C15} & {C16} & {C17} & {C18} & {C19} & {C20} & {C21} & {C22} & {C23} & {C24} \\
\midrule

86.76 & 71.03 & 80.24 & 6.70 & 0.00 & {--} & 64.68 & 72.75 & 50.16 & 0.00 & 59.77 & 21.84 & 0.00 & 99.22 \\
91.67 & 77.71 & 68.82 & 54.33 & 90.25 & {--} & 75.77 & 81.88 & 79.88 & 67.06 & 84.19 & 71.48 & 85.36 & 99.67 \\
92.18 & 70.77 & 76.33 & 93.29 & 35.87 & {--} & 67.92 & 83.22 & 79.05 & 73.07 & 84.06 & 68.77 & 67.11 & 99.96 \\
92.34 & 81.66 & 69.52 & 90.88 & 90.65 & {--} & 75.63 & 82.09 & 84.49 & 78.83 & 86.48 & 64.32 & 89.51 & 99.99 \\
92.99 & 68.44 & 55.63 & 86.61 & 68.30 & {--} & 74.85 & 88.31 & 79.52 & 76.39 & 86.57 & 69.78 & 79.21 & 99.92 \\
\rowcolor{gray!10}
93.73 & 88.14 & 73.93 & 83.54 & 94.56 & {--} & 75.56 & 87.33 & 92.86 & 86.08 & 87.83 & 83.47 & 90.07 & 99.95 \\ 
\bottomrule
\end{tabular}
\end{table}

\noindent \textbf{Evaluation on cross-geospatial tasks.} 
The motivation for this experiment arises from the fundamental constraint that labeled training data is inherently limited to specific geographic regions. 
Table~\ref{tab:cross_geospatial} reveals that DSTC~\cite{liu2024dual} fails under geospatial domain shifts, whereas a frozen DINOv2 backbone improves mIoU by +6.21\%, confirming VFM generalization. While state-of-the-art VFM-based DG semantic segmentation methods (SET~\cite{yi2024learning}, Rein~\cite{wei2024stronger}, FADA~\cite{bi2024learning}) show incremental advances, Land-MoE achieves superior performance, surpassing these benchmarks by 1.59\%, 1.99\%, and 1.77\% mIoU, respectively. These results establish Land-MoE’s capacity to overcome geographical distribution shifts for reliable land cover classification.

\begin{table}[htbp]
\centering
\caption{\textbf{Cross-geospatial Generalization Results.} Evaluation of Land-MoE vs. baselines. Table shows mAcc, mIoU, and per-class accuracy (\%) (24 classes). Land-MoE performs best overall.}
\label{tab:cross_geospatial}
\footnotesize
\setlength{\tabcolsep}{3pt} 
\renewcommand{\arraystretch}{1.1} 

\begin{tabular}{@{} l c c *{10}{c} @{}} 
\toprule
\multicolumn{1}{c}{\textbf{Method}} &
\multicolumn{2}{c}{\textbf{Overall}} &
\multicolumn{10}{c}{\textbf{Per-Class Accuracy (\%)}} \\ 
\cmidrule(lr){2-3} \cmidrule(l){4-13} 

& {mAcc} & {mIoU} & {C1} & {C2} & {C3} & {C4} & {C5} & {C6} & {C7} & {C8} & {C9} & {C10} \\ 
\midrule

DSTC\scriptsize{(ECCV'24)} & 59.71 & 46.27 & 77.36 & 76.02 & 90.39 & 75.20 & 52.44 & 93.65 & 44.97 & 54.86 & 51.81 & 27.57 \\
DINOv2\scriptsize{(Freeze)} & 69.92 & 52.48 & 82.51 & 63.11 & 86.59 & 62.87 & 68.53 & 94.51 & 26.86 & 63.40 & 72.66 & 79.88 \\
SET\scriptsize{(MM'24)} & 70.98 & 55.61 & 84.84 & 71.05 & 90.39 & 49.67 & 64.48 & 96.07 & 30.47 & 78.24 & 77.34 & 76.38 \\
Rein\scriptsize{(CVPR'24)} & 71.78 & 55.21 & 84.79 & 76.48 & 89.42 & 48.49 & 65.29 & 95.61 & 38.71 & 77.52 & 78.75 & 74.71 \\
FADA\scriptsize{(NeurIPS'24)} & 72.13 & 55.43 & 84.34 & 82.27 & 86.88 & 60.31 & 71.29 & 95.59 & 43.74 & 72.64 & 75.95 & 74.63 \\
\rowcolor{gray!10}
\textcolor{highlight}{\textbf{Land-MoE}\scriptsize{(Ours)}} & \textbf{74.18} & \textbf{57.20} & 86.99 & 75.96 & 89.05 & 68.37 & 68.37 & 95.93 & 34.53 & 82.10 & 73.52 & 79.31 \\ 
\bottomrule
\end{tabular}

\vspace{1pt} 

\setlength{\tabcolsep}{3.2pt} 
\begin{tabular}{@{} @{\hspace{4pt}} *{14}{c} @{\hspace{4pt}} @{}} 
\toprule
\multicolumn{14}{c}{\textbf{Per-Class Accuracy (\%) (Continued)}} \\
\cmidrule(l){1-14} 

{C11} & {C12} & {C13} & {C14} & {C15} & {C16} & {C17} & {C18} & {C19} & {C20} & {C21} & {C22} & {C23} & {C24} \\ 
\midrule

62.68 & 88.62 & 82.61 & 46.19 & 73.81 & 7.01 & 92.93 & 85.48 & 42.14 & 9.62 & 69.96 & 53.14 & 42.98 & 31.51 \\
64.82 & 88.05 & 90.26 & 66.61 & 68.36 & 13.15 & 96.28 & 91.81 & 52.18 & 37.31 & 78.33 & 72.53 & 75.56 & 81.84 \\
57.43 & 88.16 & 94.79 & 53.14 & 79.45 & 3.94 & 97.41 & 92.11 & 75.31 & 40.40 & 80.57 & 74.44 & 68.56 & 78.80 \\
63.43 & 89.01 & 94.66 & 57.02 & 75.66 & 5.58 & 96.98 & 91.61 & 70.48 & 52.36 & 81.72 & 76.05 & 72.96 & 65.54 \\
72.84 & 89.35 & 85.54 & 56.39 & 31.86 & 29.33 & 96.18 & 91.19 & 70.02 & 48.51 & 79.49 & 76.97 & 77.03 & 78.68 \\
\rowcolor{gray!10}
63.94 & 86.79 & 90.37 & 61.20 & 77.96 & 14.52 & 97.42 & 90.63 & 68.72 & 53.97 & 81.70 & 77.50 & 73.16 & 88.39 \\ 
\bottomrule
\end{tabular}
\end{table}

\subsection{Ablation Studies}
\label{exp:ablation_study}

\noindent \textbf{Analysis of key components in Land-MoE.}
We analyze the contribution of Land-MoE's key components, the MoLTE and FAF modules, as presented in Table~\ref{tab:ablation}. The baseline configuration, employing only a Mask2Former decoder with a frozen VFM backbone, yields suboptimal performance.
The FAF module alone demonstrates substantial improvements, enhancing mIoU by 6.78\% (cross-sensor) and 3.38\% (cross-geospatial). Similarly, the MoLTE module achieves mIoU gains of 8.47\% and 4.02\% respectively.
The full Land-MoE framework, which integrates both modules, attains peak performance, establishing that each component contributes distinct and complementary capabilities, which underscores the architectural rationale behind Land-MoE's design.

\begin{table}[htbp]
\centering
\caption{\textbf{Land-MoE Ablation Study.} Cross-sensor/geospatial performance results (Params*, mAcc, mIoU) for configurations with frozen DINOv2 + Mask2Former decoder, varying MoLTE/FAF ($\checkmark$ used). Params*: trainable PEFT (excluding 20.6M fixed decoder). Full Land-MoE shows highest accuracy.}
\label{tab:ablation}
\setlength{\tabcolsep}{3.5pt}
\begin{tabular}{@{}ccc*{3}{>{\centering\arraybackslash}m{11mm}}*{3}{>{\centering\arraybackslash}m{11mm}}@{}} 
\toprule
\multicolumn{3}{c}{Components} & 
\multicolumn{3}{c}{Cross-sensor} & 
\multicolumn{3}{c}{Cross-geospatial} \\
\cmidrule(lr){1-3} \cmidrule(lr){4-6} \cmidrule(lr){7-9}
VFM & MoLTE & FAF & Params\textsuperscript{*} & mAcc & mIoU & Params\textsuperscript{*} & mAcc & mIoU \\
\midrule

$\checkmark$ & \ding{55} & \ding{55} 
& 0.00M & 69.10 & 53.37 
& 0.00M & 69.92 & 52.48 \\

$\checkmark$ & \ding{55} & $\checkmark$ 
& 0.64M & 74.63 & 60.15 
& 0.64M & 72.34 & 55.86 \\

$\checkmark$ & $\checkmark$ & \ding{55} 
& 3.71M & 75.15 & 61.84 
& 2.25M & 73.00 & 56.50 \\

\rowcolor{gray!10}
$\checkmark$ & $\checkmark$ & $\checkmark$ 
& 3.81M & \textbf{77.95} & \textbf{64.36} 
& 2.89M & \textbf{74.18} & \textbf{57.20} \\
\bottomrule
\end{tabular}
\end{table}

\noindent \textbf{Evaluation of different VFMs.}
To assess Land-MoE's adaptability across various VFMs, we evaluate its performance alongside baseline methods under diverse VFM configurations on cross-sensor and cross-geospatial MLCC tasks, as detailed in Table~\ref{tab:diff_VFMS}. Experiments are conducted using CLIP~\cite{radford2021learning}, SAM~\cite{kirillov2023segment}, EVA02~\cite{fang2024eva}, and DINOv2~\cite{oquab2023dinov2} as VFM backbones. For a fair comparison, all methods employ the Mask2Former decoder; the trainable parameters reported in the table quantify only PEFT modules. 
Across all evaluated VFM backbones, the VFM-based DG semantic segmentation methods (SET, Rein, FADA) consistently achieve better performance than the baseline strategy of freezing the VFM and training only the decoder. Importantly, our proposed Land-MoE consistently surpasses all baselines when using CLIP, SAM, EVA02, or DINOv2 as the VFM, highlighting its broad compatibility.

\begin{table}[t]
\centering
\caption{\textbf{VFM and PEFT Method Comparison.} Cross-sensor/geospatial performance results (mAcc, mIoU, Params*). Table shows Land-MoE vs. baselines across various VFMs. Params*: trainable PEFT (excluding 20.6M fixed decoder). Land-MoE consistently highest.}
\label{tab:diff_VFMS} 
\small
\setlength{\tabcolsep}{4pt} 
\begin{tabular}{@{}c c *{3}{>{\centering\arraybackslash}m{11mm}} *{3}{>{\centering\arraybackslash}m{11mm}}@{}}
\toprule
\multirow{2}{*}{\textbf{VFM}} & 
\multirow{2}{*}{\textbf{PEFT Methods}} & 
\multicolumn{3}{c}{\textbf{Cross-sensor}} & 
\multicolumn{3}{c}{\textbf{Cross-geospatial}} \\ 
\cmidrule(lr){3-5} \cmidrule(lr){6-8} 
& & 
Params\textsuperscript{*} & mAcc & mIoU & 
Params\textsuperscript{*} & mAcc & mIoU \\ 
\midrule

\multirow{5}{*}{CLIP (Large)~\cite{radford2021learning}} 
 & Freeze & 0.00M & 61.59 & 46.30 & 0.00M & 64.51 & 48.10 \\
 & SET~\cite{yi2024learning} & 7.55M & 60.93 & 48.18 & 7.55M & 62.56 & 49.26 \\
 & Rein~\cite{wei2024stronger} & 2.99M & 69.24 & 56.70 & 2.99M & 69.95 & 53.37 \\
 & FADA~\cite{bi2024learning} & 2.06M & 68.68 & 55.03 & 2.06M & 71.06 & 51.86 \\
 \rowcolor{gray!10}
 & Land-MoE & 3.81M & \textbf{73.25} & \textbf{61.94} & 2.89M & \textbf{71.16} & \textbf{54.71} \\
\midrule

\multirow{5}{*}{SAM (Huge)~\cite{kirillov2023segment}}
 & Freeze & 0.00M & 59.35 & 44.98 & 0.00M & 58.96 & 44.63 \\
 & SET~\cite{yi2024learning} & 7.68M & 65.15 & 50.31 & 7.68M & 64.75 & 52.10 \\
 & Rein~\cite{wei2024stronger} & 3.89M & 60.72 & 46.13 & 3.89M & 69.42 & 50.27 \\
 & FADA~\cite{bi2024learning} & 2.45M & 63.37 & 46.39 & 2.45M & 68.25 & 50.04 \\
 \rowcolor{gray!10}
 & Land-MoE & 3.30M & \textbf{72.54} & \textbf{60.57} & 3.12M & \textbf{69.95} & \textbf{52.91} \\
\midrule

\multirow{5}{*}{EVA02 (Large)~\cite{fang2024eva}}
 & Freeze & 0.00M & 59.11 & 45.48 & 0.00M & 61.41 & 49.84 \\
 & SET~\cite{yi2024learning} & 7.55M & 56.56 & 46.29 & 7.55M & 66.69 & 51.57 \\
 & Rein~\cite{wei2024stronger} & 2.99M & 63.82 & 50.81 & 2.99M & 69.62 & 51.33 \\
 & FADA~\cite{bi2024learning} & 2.06M & 62.24 & 46.49 & 2.06M & 69.09 & 51.27 \\
 \rowcolor{gray!10}
 & Land-MoE & 3.81M & \textbf{72.76} & \textbf{59.85} & 2.89M & \textbf{71.45} & \textbf{53.87} \\
\midrule

\multirow{5}{*}{DINOv2 (Large)~\cite{oquab2023dinov2}}
 & Freeze & 0.00M & 69.10 & 53.37 & 0.00M & 69.92 & 52.48 \\
 & SET~\cite{yi2024learning} & 7.55M & 70.24 & 55.73 & 7.55M & 70.98 & 55.61 \\
 & Rein~\cite{wei2024stronger} & 2.99M & 73.44 & 59.06 & 2.99M & 71.78 & 55.21 \\
 & FADA~\cite{bi2024learning} & 2.06M & 73.21 & 56.57 & 2.06M & 72.13 & 55.43 \\
 \rowcolor{gray!10}
 & Land-MoE & 3.81M & \textbf{77.95} & \textbf{64.36} & 2.89M & \textbf{74.18} & \textbf{57.20} \\
\bottomrule
\end{tabular}
\end{table}





\section{Conclusion}
We introduce Land-MoE, a novel approach for large-scale cross-scene multispectral land cover classification that effectively mitigates spectral shifts between source and target domains. Land-MoE efficiently fine-tunes Vision Foundation Models using its Frequency-aware Mixture of Low-rank Token Experts as adapters to achieve strong cross-domain generalization. Extensive experiments across various sensors and geographical regions demonstrate that Land-MoE achieves state-of-the-art performance in large-scale cross-scene multispectral land cover classification tasks and also demonstrates strong performance in RGB remote sensing image domain generalization semantic segmentation. Code is provided in the supplementary material for review and will be available upon publication.

\noindent \textbf{Limitation.}
\label{sec:limitaion}
Although Land-MoE demonstrates robust performance in multispectral and RGB-based land cover classification via  VFMs, its current architecture faces limitations in processing hyperspectral remote sensing imagery due to the high dimensionality of spectral bands.


\noindent \textbf{Broader impact.}
\label{sec:broder_impact}
Land-MoE enables high-precision, data-efficient land cover classification across diverse conditions, improving global environmental monitoring, urban planning, and resource management accessibility. However, its precision may raise privacy concerns.





\newpage

\bibliographystyle{abbrvnat}  
\bibliography{main}

\appendix
\section{Details of the Construction of Cross-Sensor and Cross-Geospatial Generalization Tasks}
\label{sec:detail_dataset}
\subsection{Data Sources}
Our experiments utilize MSIs acquired by four distinct satellite platforms: GF-2, PlanetScope, GF-1, and Sentinel-2. GF-2, part of the High-Definition Earth observation (HDEOS) program by CNSA, captures data in four bands: blue (0.45--0.52 $\mu$m), green (0.52--0.59 $\mu$m), red (0.63--0.69 $\mu$m), and near-infrared (0.77--0.89 $\mu$m), with a nominal spatial resolution of 4 m. PlanetScope, operated by Planet Labs, acquires imagery in four spectral bands: blue (0.46--0.52 $\mu$m), green (0.50--0.59 $\mu$m), red (0.59--0.67 $\mu$m), and near-infrared (0.78--0.86 $\mu$m), with a spatial resolution varying between 3.7 m and 4.1 m. GF-1, as the first satellite of the HDEOS program, carries a multispectral sensor that captures the same four bands as GF-2 but at a coarser spatial resolution of 8 m. Finally, from the European Union's Copernicus programme, we incorporate Sentinel-2 data, specifically selecting the 10 m-resolution bands corresponding to blue (central wavelength 0.49 $\mu$m, Band 2), green (central wavelength 0.56 $\mu$m, Band 3), red (central wavelength 0.66 $\mu$m, Band 4), and near-infrared (central wavelength 0.83 $\mu$m, Band 8). For consistent processing and model input, all images were uniformly cropped to a spatial dimension of $512 \times 512$ pixels.
\begin{figure}[htbp]
    \centering
	\includegraphics[ width=1.0\linewidth]{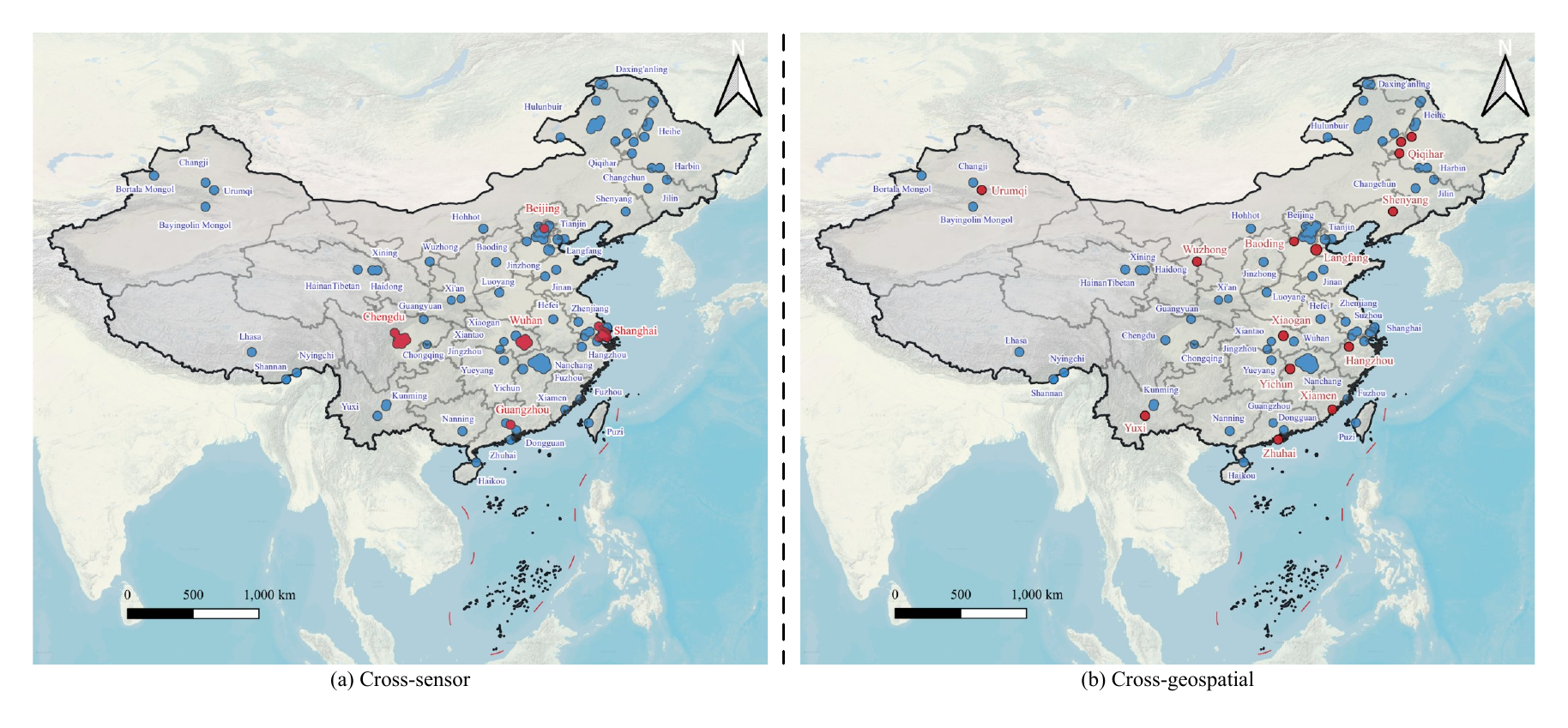}
	\caption{\textbf{Geographical distribution of SD and TDs for the constructed cross-sensor and cross-geospatial generalization tasks.} Subfigure (a) presents the domain distribution for the cross-sensor task, where locations corresponding to the SD (GF-2 imagery) are marked by blue solid circles, and those corresponding to the TDs (PlanetScope, GF-1, and Sentinel-2 imagery) are indicated by red circles. Subfigure (b) illustrates the domain distribution for the cross-geospatial task, with blue solid circles representing the SD (GF-2 imagery from various regions) and red solid circles denoting the TD (GF-2 imagery from designated cities).}
    \label{fig:Task_construct}
\end{figure}
\subsection{Cross-Sensor Generalization Task}
For the cross-sensor generalization task, GF-2 MSIs from the Five-Billion-Pixels~\cite{tong2023enabling} dataset are designated as the SD training data. In parallel, MSIs acquired by PlanetScope, GF-1, and Sentinel-2 serve as the TDs to evaluate model generalization performance specifically under sensor shifts. As illustrated in Figure~\ref{fig:Task_construct}(a), the geographical distribution of the SD (GF-2 imagery) is marked by blue solid circles, whereas the TD data (PlanetScope, GF-1, and Sentinel-2) are indicated by red circles. Specifically, the PlanetScope TD data covers the cities of Chengdu and Shanghai, the GF-1 TD data is sourced from Wuhan, and the Sentinel-2 TD data is collected from Beijing and Guangzhou.

\subsection{Cross-Geospatial Generalization Task}
The cross-geospatial generalization task is established utilizing a subset of 150 GF-2 MSIs sourced from 62 distinct administrative regions across China, as provided in the Five-Billion-Pixels~\cite{tong2023enabling} dataset. We partition these images based on their geographical origin to define the SD and TD. Images originating from the administrative regions of Yichun, Baoding, Xiaogan, Langfang, Yuxi, Qiqihar, Hangzhou, Zhuhai, Urumqi, Wuzhong, Xiamen, and Shenyang are collectively designated as the TD. Their geographical distributions are indicated by solid red circles in Figure~\ref{fig:Task_construct}(b). The remaining MSIs, corresponding to administrative regions geographically distinct from the TD locations, are utilized as the SD training data. Their distributions are represented by solid blue circles in Figure~\ref{fig:Task_construct}(b).

\subsection{Classification System}
\label{sec:sbec_cs}
Our land cover classification system comprises the 24 distinct categories defined within the Five-Billion-Pixels~\cite{tong2023enabling} dataset. These categories and their corresponding codes are as follows: C1: Industrial areas, C2: Paddy fields, C3: Irrigated fields, C4: Dry farmland, C5: Vegetable plots, C6: Arbor forests, C7: Shrub forests, C8: Parks, C9: Natural grasslands, C10: Artificial grasslands, C11: Rivers, C12: Urban residential areas, C13: Lakes, C14: Ponds, C15: Fish ponds, C16: Snow cover, C17: Bare land, C18: Rural residential areas, C19: Stadiums, C20: Squares, C21: Roads, C22: Interchanges, C23: Railway stations, and C24: Airports.
\section{Additional Parameter Analysis}
\label{sec:detail_param_analysis}

\subsection{Effects of Learning Rate and Batch Size}
We investigated the impact of two key optimization hyperparameters, the learning rate and batch size, on the performance and generalization capability of Land-MoE.

Table~\ref{tab:learning_rate_comparison} illustrates the model's performance under different learning rate configurations, specifically examining values in the range of $[5\times 10^{-4}, 1\times 10^{-4}, 5\times 10^{-5}, 1\times 10^{-5}, 1\times 10^{-6}]$ on both the cross-sensor and cross-geospatial generalization tasks. The experimental results clearly demonstrate that Land-MoE achieves the highest mAcc and mIoU metrics on both tasks when the learning rate is set to $1\times 10^{-4}$. Further analysis suggests that excessively large learning rates (e.g., $5\times 10^{-4}$) may lead to optimization instability, potentially disrupting the fine-tuning process and compromising the retention of useful features learned by the pre-trained model. Conversely, overly small learning rates (e.g., $1\times 10^{-5}, 1\times 10^{-6}$) can impede effective parameter updates, thereby limiting the model’s capacity to adapt sufficiently to cross-scenario variations and resulting in suboptimal performance.

Table~\ref{tab:batch_size_comparison} further examines the influence of varying batch size configurations on Land-MoE’s generalization capability across the same tasks. The results indicate that the model attains optimal performance on both cross-sensor and cross-geospatial generalization tasks with a batch size of 8. This suggests that a batch size of 8 strikes an optimal balance between the stability of the training process (e.g., gradient estimation) and the model's ability to generalize effectively to unseen TDs.


\begin{table}[htbp] 
\centering
\caption{\textbf{Effects of learning rate on Land-MoE performance for cross-sensor and cross-geospatial generalization.} The table presents mAcc and mIoU metrics for learning rates ranging from $5\times 10^{-4}$ to $1\times 10^{-6}$. Optimal values for each metric are indicated in bold.}
\begin{tabular}{llccccc} 
\toprule
\multirow{2}{*}{Task} & \multirow{2}{*}{Metric} & \multicolumn{5}{c}{Learning Rate} \\
\cmidrule(lr){3-7} 
                      &                         & 5e-4  & 1e-4           & 5e-5  & 1e-5  & 1e-6  \\
\midrule
\multirow{2}{*}{Cross-sensor} & mAcc & 19.41 & \textbf{77.95} & 76.72 & 65.12 & 27.49 \\
\cmidrule(lr){2-7} 
                              & mIoU & 10.97 & \textbf{64.36} & 62.73 & 48.13 & 18.16 \\
\midrule 
\multirow{2}{*}{Cross-geospatial} & mAcc & 22.52 & \textbf{74.18} & 72.00 & 64.57 & 35.39 \\
\cmidrule(lr){2-7} 
                                & mIoU & 14.75 & \textbf{57.20} & 54.92 & 47.65 & 24.85 \\
\bottomrule
\end{tabular}
\label{tab:learning_rate_comparison}
\end{table}


\begin{table}[htbp] 
\centering
\caption{\textbf{Effects of batch size on Land-MoE performance for cross-sensor and cross-geospatial generalization tasks.} The table presents mAcc and mIoU metrics for batch sizes 4, 8, and 16. Optimal values for each metric are indicated in bold.}
\begin{tabular}{llccc} 
\toprule
\multirow{2}{*}{Task} & \multirow{2}{*}{Metric} & \multicolumn{3}{c}{Batch size} \\
\cmidrule(lr){3-5} 
                      &                         & 4     & 8              & 16    \\
\midrule
\multirow{2}{*}{Cross-sensor} & mAcc & 75.44 & \textbf{77.95} & 75.99 \\
\cmidrule(lr){2-5} 
                              & mIoU & 60.68 & \textbf{64.36} & 63.54 \\
\midrule 
\multirow{2}{*}{Cross-geospatial} & mAcc & 70.86 & \textbf{74.18} & 72.67 \\
\cmidrule(lr){2-5} 
                                & mIoU & 53.11 & \textbf{57.20} & 56.86 \\
\bottomrule
\end{tabular}
\label{tab:batch_size_comparison} 
\end{table}

\subsection{Analysis of the Number of Learnable Low-Rank Token Experts and Low-Rank Dimensions}
The MoLTE component in Land-MoE is designed to enhance robustness to spectral shifts and enable instance-specific adaptation through rank-diversified learnable low-rank tokens. We investigated the impact of the number of learnable low-rank token experts ($N_e$) and their corresponding low-rank dimensions ($r_k$) on cross-scene MLCC performance. Table~\ref{tab:config} summarizes the results for various configurations. For the cross-sensor generalization task, Land-MoE achieves the highest mAcc and mIoU when utilizing $N_e=3$ experts with heterogeneous low-rank dimensions $r_k \in \{8, 16, 32\}$. For the cross-geospatial generalization task, optimal mAcc and mIoU are attained with $N_e=2$ experts and low-rank dimensions $r_k \in \{8, 16\}$. These results highlight the task-specific sensitivity to the configuration of the expert layer.

\begin{table}[htbp]
\centering
\caption{\textbf{Analysis of the impact of the number of learnable low-rank token experts ($N_e$) and their corresponding low-rank dimensions ($r_k$).} The table presents mAcc and mIoU for cross-sensor and cross-geospatial generalization tasks across various $\{N_e, r_k\}$ configurations, along with the respective number of trainable parameters\textsuperscript{*}. Optimal values for each task and metric are indicated in bold.}
\label{tab:config}
\setlength{\tabcolsep}{3.5pt} 
\begin{tabular}{@{}l@{\hspace{6pt}} 
>{\centering\arraybackslash}m{14mm}@{\hspace{4pt}} 
*{2}{>{\centering\arraybackslash}m{14mm}@{\hspace{2pt}}} 
*{2}{>{\centering\arraybackslash}m{14mm}@{}}} 
\toprule
\multirow{2}{*}{Method} & 
\multirow{2}{*}{\makecell{Trainable\\ Params\textsuperscript{*}}} &
\multicolumn{2}{c}{Cross-sensor} & 
\multicolumn{2}{c}{Cross-geospatial} \\ 
\cmidrule(lr){3-4} \cmidrule(l){5-6}
& & mAcc & mIoU & mAcc & mIoU \\ 
\midrule

$N_e=2, r_k \in \{8,16\}$ & 2.89M & 76.33 & 64.05 & \textbf{74.18} & \textbf{57.20} \\
$N_e=3, r_k \in \{8,16,32\}$ & 3.81M & \textbf{77.95} & \textbf{64.36} & 73.93 & 56.27 \\
$N_e=4, r_k \in \{8,16,32,48\}$ & 5.15M & 76.46 & 63.46 & 73.26 & 56.15 \\
$N_e=5, r_k \in \{8,16,32,48,64\}$ & 6.93M & 76.81 & 64.15 & 73.38 & 56.05 \\
$N_e=6, r_k \in \{8,16,32,48,64,96\}$ & 9.56M & 75.77 & 63.24 & 73.52 & 56.14 \\
\bottomrule
\end{tabular}
\end{table}

\subsection{Analysis of the Sequence Length of Learnable Tokens per Expert}
We analyzed the impact of the sequence length ($m$) of learnable low-rank tokens within each expert on the overall model performance. Table~\ref{tab:token_analysis} presents the results for sequence lengths ranging from 50 to 200. For both cross-sensor and cross-geospatial generalization tasks, the optimal performance is consistently achieved with a token sequence length of $m=100$. While shorter sequence lengths would reduce the number of trainable parameters within the experts, our experiments demonstrate that a sequence length of 100 for the learnable tokens within Land-MoE is necessary to ensure the highest accuracy in large-scale cross-scene MLCC tasks, suggesting this length provides sufficient representational capacity.

\begin{table}[ht]
\centering
\caption{\textbf{Analysis of the impact of the sequence length ($m$) of learnable low-rank tokens within each expert.} The table presents mAcc, mIoU, and trainable parameters\textsuperscript{*} for cross-sensor and cross-geospatial generalization tasks across varying token lengths. Optimal values for each task and metric are indicated in bold.}
\label{tab:token_analysis}
\begin{tabular}{@{}llccccccc@{}}
\toprule
\multirow{2}{*}{Tasks} & \multirow{2}{*}{Metrics} & \multicolumn{7}{c}{Token Length} \\
\cmidrule(lr){3-9}
 & & 50 & 75 & 100 & 125 & 150 & 175 & 200 \\
\midrule
\multirow{3}{*}{Cross-sensor} 
 & mAcc (\%) & 76.29 & 77.08 & \textbf{77.95} & 76.55 & 77.08 & 76.30 & 73.31 \\
 & mIoU (\%) & 63.16 & 63.85 & \textbf{64.36} & 63.30 & 64.31 & 62.52 & 59.17 \\
 & Params\textsuperscript{*} (M) & 3.74 & 3.77 & 3.81 & 3.84 & 3.87 & 3.91 & 3.94 \\
\midrule
\multirow{3}{*}{Cross-geospatial}
 & mAcc (\%) & 73.06 & 73.76 & \textbf{74.18} & 73.70 & 73.29 & 73.66 & 73.98 \\
 & mIoU (\%) & 56.31 & 56.95 & \textbf{57.20} & 55.67 & 55.90 & 55.91 & 55.75 \\
 & Params\textsuperscript{*} (M) & 2.87 & 2.88 & 2.89 & 2.91 & 2.92 & 2.94 & 2.95 \\
\bottomrule
\end{tabular}
\end{table}

\subsection{Effects of Different Embedding Positions of Land-MoE within VFMs}
To investigate how the placement of Land-MoE modules within the layers of a VFM affects cross-scene generalization performance, we designed five comparative experiments based on the DINOv2-Large model, which features a 24-layer Vision Transformer (ViT) architecture. The embedding strategies explored were: \textbf{Freeze}, where the pre-trained VFM is used without any fine-tuning of its parameters and without adding Land-MoE; \textbf{Shallow}, where Land-MoE modules are embedded only after the first 6 layers (shallow layers) of the VFM; \textbf{Deep}, where Land-MoE modules are embedded only after the last 6 layers (deep layers); \textbf{Specific}, where Land-MoE modules are strategically embedded only after specific layers (7, 11, 15, and 23), corresponding to the feature layers typically connected to the Mask2Former decoder; and \textbf{Land-MoE (Proposed)}, where Land-MoE modules are added after each ViT Block in every layer of the VFM, enabling comprehensive layer-wise feature refinement. Table~\ref{tab:method_layer_comparison} summarizes the results of MLCC for each of these strategies on both cross-sensor and cross-geospatial generalization tasks. The experimental results clearly indicate that deploying Land-MoE in all layers and refining VFM features layer-by-layer (Strategy 5, the proposed method) significantly improves the model's cross-scene classification performance compared to other placement strategies. This strongly validates the effectiveness of the proposed layer-by-layer adaptive adjustment of VFM features facilitated by Land-MoE.

\begin{table}[htbp]
\centering
\caption{\textbf{Analysis of the impact of different Land-MoE embedding positions within the VFM layers.} The table presents mAcc and mIoU for cross-sensor and cross-geospatial generalization tasks across various embedding strategies (Freeze, Shallow, Deep, Specific, Land-MoE), indicating the layers where modules are applied. Optimal performance is highlighted in bold.}
\label{tab:method_layer_comparison}
\setlength{\tabcolsep}{4pt}
\begin{tabular}{@{}l@{\hspace{6pt}} 
>{\centering\arraybackslash}m{26mm}@{\hspace{4pt}} 
*{2}{>{\centering\arraybackslash}m{14mm}@{\hspace{2pt}}} 
*{2}{>{\centering\arraybackslash}m{14mm}@{}}} 
\toprule
\multirow{2}{*}{Method} & 
\multirow{2}{*}{Layer} & 
\multicolumn{2}{c}{Cross-sensor} & 
\multicolumn{2}{c}{Cross-geospatial} \\
\cmidrule(lr){3-4} \cmidrule(l){5-6}
& & mAcc & mIoU & mAcc & mIoU \\
\midrule
Freeze    & None        & 69.10 & 53.37 & 69.92 & 52.48 \\
Shallow   & [0,1,2,3,4,5] & 76.47 & 63.60 & 73.99 & 56.66 \\
Deep      & [18,19,20,21,22,23] & 75.46 & 60.70 & 71.70 & 55.56 \\
Specific  & [7, 11, 15, 23] & 76.50 & 62.91 & 73.26 & 55.61 \\
\rowcolor{gray!10}
Land-MoE    & Full        & \textbf{77.95} & \textbf{64.36} & \textbf{74.18} & \textbf{57.20} \\
\bottomrule
\end{tabular}
\end{table}

\section{Generalization Performance of Land-MoE on Natural Remote Sensing Images}
\label{sec:performance_RGB_generalization}

\subsection{Cross-Scene Task Construction}
Although Land-MoE is primarily designed for large-scale cross-scene MLCC tasks, it is also evaluated for its adaptability in handling domain shift issues in natural remote sensing images (RGB). To validate the effectiveness of Land-MoE in the context of domain generalization on natural remote sensing images, we constructed two distinct cross-scene classification tasks.

We constructed two distinct cross-scene land cover classification tasks for natural remote sensing images. The first task, denoted as \textbf{Rural2Urban}, is established based on the LoveDA dataset~\cite{wang2021loveda}, where the rural scene portion serves as the SD and the urban scene constitutes the TD. The second task, denoted as \textbf{Potsdam2Vaihingen}, utilizes two very-high-resolution true orthophoto natural remote sensing datasets: Potsdam (specifically its R, G, B bands) and Vaihingen (Nir, R G bands). In this task, the Potsdam dataset is designated as the SD, while the Vaihingen dataset serves as the TD.

In all experiments for these tasks, images were uniformly cropped to a spatial dimension of $512 \times 512$ pixels to ensure consistency during the training and evaluation processes.

\subsection{Experimental Results and Analysis}
We present a performance evaluation of Land-MoE and compare it against several existing state-of-the-art methods on the constructed natural remote sensing image cross-scene land cover classification tasks. Tables~\ref{tab:loveda_results} and~\ref{tab:vaihingen_results} summarize the comparative results for the Rural2Urban and Potsdam2Vaihingen tasks, respectively.

In the Rural2Urban task (Table~\ref{tab:loveda_results}), Land-MoE demonstrates strong performance. Specifically, it achieves a notable $9.94\%$ improvement in mIoU compared to DSTC~\cite{liu2024dual}, a leading method for MLCC. Furthermore, Land-MoE shows an $3.62\%$ improvement in mIoU over a baseline method that utilizes a frozen DINOv2 backbone with only the Mask2Former decoder trained. Compared to Rein~\cite{wei2024stronger}, a semantic segmentation method known for its DG performance, Land-MoE exhibits a $1.40\%$ improvement in mIoU.

On the Potsdam2Vaihingen task (Table~\ref{tab:vaihingen_results}), Land-MoE similarly exhibits a significant advantage. We note the particularly low performance of DSTC on this task. This is primarily attributed to the substantial spectral shift between the SD (Potsdam, utilizing R, G, B bands) and the TD (Vaihingen, utilizing Nir, R, G bands). Despite this challenge, Land-MoE outperforms DSTC by $44.55\%$ in mIoU. Land-MoE also demonstrates superiority over the frozen DINOv2 + Mask2Former decoder baseline, achieving a $5.78\%$ mIoU improvement, and over Rein, with a $1.83\%$ mIoU improvement.

Overall, the results on both natural remote sensing image cross-scene tasks validate Land-MoE's robust land cover classification capabilities and its excellent performance in mitigating domain shifts, even when applied to data types beyond its primary multispectral focus.

\begin{table}[t] 
\centering
\caption{\textbf{Performance evaluation of Land-MoE and state-of-the-art methods on the Rural2Urban cross-scene semantic segmentation task using the LoveDA dataset.} The table reports overall mAcc and mIoU, as well as per-class accuracy. Optimal values are highlighted in bold. Refer to the note below the table for class abbreviations.} 
\label{tab:loveda_results} 
\setlength{\tabcolsep}{6.5pt} 
\renewcommand{\arraystretch}{1.1} 

\begin{tabular}{@{} l c c *{7}{c} @{}} 
\toprule
\multicolumn{1}{c}{\textbf{Method}} &
\multicolumn{2}{c}{\textbf{Overall}} &
\multicolumn{7}{c}{\textbf{Per-Class Accuracy (\%)}} \\ 
\cmidrule(lr){2-3} \cmidrule(l){4-10} 

& {mAcc} & {mIoU} &
{BG} & {BU} & {RD} & {WT} & {BR} & {FR} & {AG} \\ 
\midrule

DSTC\scriptsize{(ECCV'24)} & 62.90 & 47.78 & 61.84 & 66.04 & 60.44 & 78.40 & 51.42 & 67.83 & 54.35 \\
DINOv2\scriptsize{(Freeze)} & 68.39 & 54.10 & 62.75 & 80.04 & 69.38 & 80.51 & 56.88 & 62.66 & 66.50 \\
SET\scriptsize{(ACM MM'24)} & 71.36 & 54.69 & 58.74 & 82.22 & 73.29 & 80.32 & 65.82 & 73.95 & 65.21 \\
Rein\scriptsize{(CVPR'24)} & 71.70 & 56.32 & 60.82 & 79.52 & 75.29 & 81.72 & 59.93 & 77.01 & 67.60 \\
FADA\scriptsize{(NeurIPS'24)} & 71.68 & 56.33 & 64.45 & 82.87 & 74.91 & 83.25 & 61.74 & 74.42 & 60.10 \\
\rowcolor{gray!10} 
\textbf{Land-MoE\scriptsize{(Ours)}} & \textbf{73.99} & \textbf{57.72} & 63.36 & 81.78 & 74.48 & 83.58 & 77.58 & 73.22 & 63.95 \\ 

\bottomrule 
\end{tabular}

\vspace{4pt} 
\raggedright 
\scriptsize \textit{Note}: All values in \%. Abbreviations: BG=Background, BU=Building, RD=Road, WT=Water, BR=Barren, FR=Forest, AG=Agricultural. Bold indicates best results. 

\end{table}

\begin{table}[htbp] 
\centering
\caption{\textbf{This table presents the performance evaluation of Land-MoE and state-of-the-art methods on the challenging Potsdam2Vaihingen cross-scene land cover classification task.} This task is characterized by a significant spectral shift (Potsdam RGB $\rightarrow$ Vaihingen NirRG). Metrics include overall mAcc and mIoU, in addition to per-class accuracy. Optimal values are highlighted in bold. Refer to the note below the table for class abbreviations.}
\label{tab:vaihingen_results} 
\setlength{\tabcolsep}{5.5pt} 
\renewcommand{\arraystretch}{1.1} 

\begin{tabular}{@{} l c c *{6}{c} @{}} 
\toprule
\multicolumn{1}{c}{\textbf{Method}} &
\multicolumn{2}{c}{\textbf{Overall}} &
\multicolumn{6}{c}{\textbf{Per-Class Accuracy (\%)}} \\ 
\cmidrule(lr){2-3} \cmidrule(l){4-9} 

& {mAcc} & {mIoU} &
{Imp. Surf.} & {Build.} & {Low Veg.} & {Tree} & {Car} & {Clutter} \\ 
\midrule
DSTC\scriptsize{(ECCV'24)} & 32.54 & 16.62 & 56.74 & 52.69 & 3.90 & 0.23 & 1.27 & 80.38 \\
DINOv2\scriptsize{(Freeze)} & 77.68 & 55.39 & 79.62 & 93.65 & 58.25 & 74.86 & 69.79 & 88.93 \\
SET\scriptsize{(ACM MM'24)} & 78.65 & 56.08 & 77.48 & 96.23 & 49.66 & 83.85 & 69.30 & 95.35 \\
Rein\scriptsize{(CVPR'24)} & 80.53 & 59.34 & 80.40 & 95.87 & 61.27 & 85.94 & 64.84 & 94.84 \\
FADA\scriptsize{(NeurIPS'24)} & 81.12 & 59.27 & 83.08 & 94.85 & 54.09 & 89.57 & 71.22 & 93.94 \\
\rowcolor{gray!10} 
\textbf{Land-MoE\scriptsize{(Ours)}} & \textbf{81.98} & \textbf{61.17} & 80.66 & 94.72 & 65.08 & 84.86 & 73.29 & 93.28 \\ 

\bottomrule 
\end{tabular}

\vspace{4pt} 
\raggedright 
\scriptsize \textit{Note}: All values are in \%. Abbreviations: Imp. Surf.=Impervious surface, Build.=Building, Low Veg.=Low vegetation. Bold indicates best results. 

\end{table}

\section{Additional Cross-Scene Land Cover Classification Results}
\label{sec:additional_results}

To complement the quantitative analysis presented in the main paper and preceding sections of the supplementary material, we provide additional qualitative results showcasing the performance of Land-MoE and compared methods on the constructed cross-scene generalization tasks.

\begin{figure}[htbp]
    \centering
	\includegraphics[ width=1.0\linewidth]{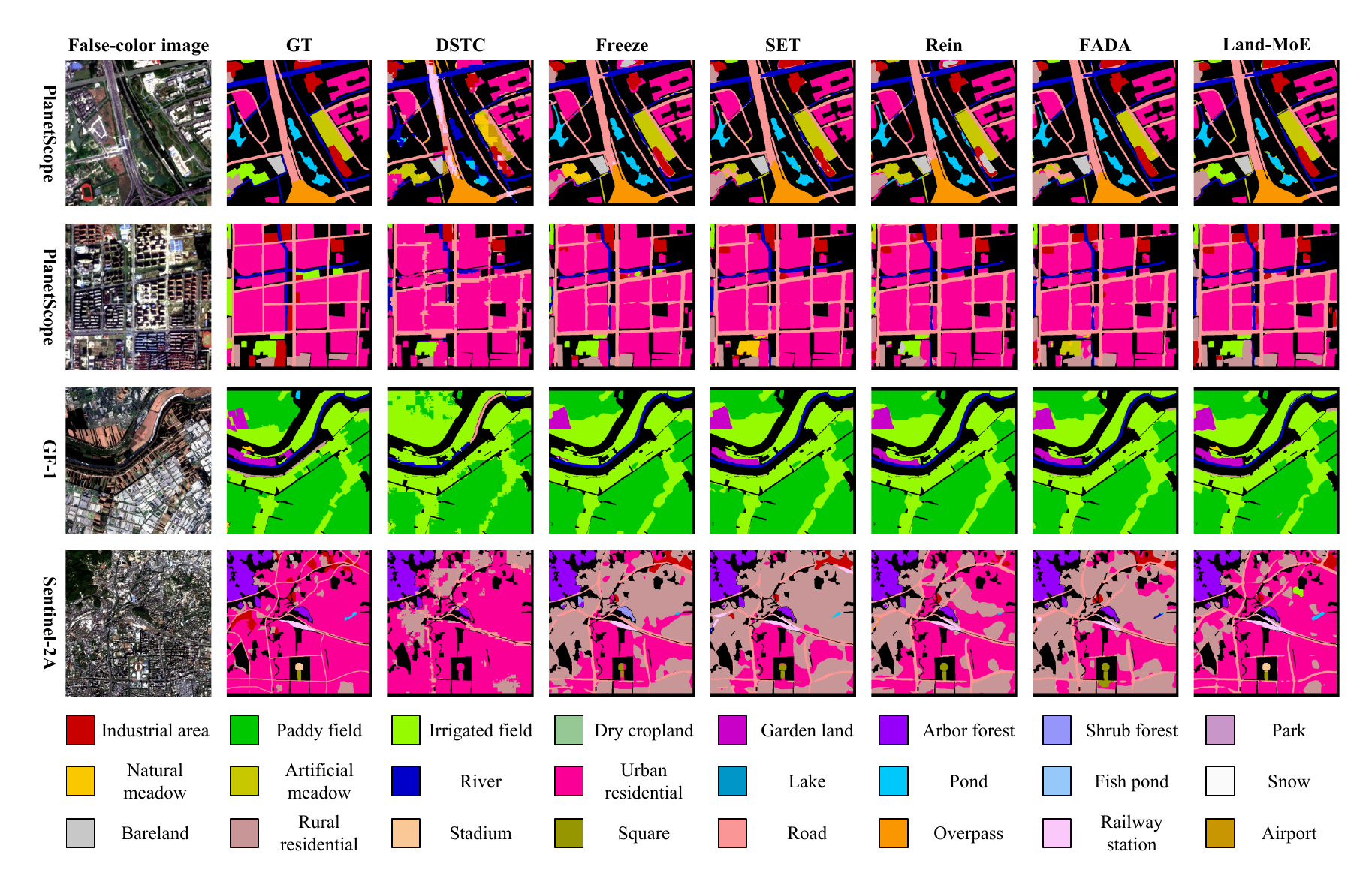}
        \caption{\textbf{Qualitative results showing predicted land cover classification maps for the cross-sensor generalization task.} The figure illustrates the performance of Land-MoE in comparison to state-of-the-art baseline methods on cross-scene multispectral remote sensing images.}
    \label{fig:Cross_sensor_more}
\end{figure}

Figure~\ref{fig:Cross_sensor_more} presents additional visual examples of predicted land cover classification maps for cross-scene MLCC within the cross-sensor generalization task, illustrating the performance of Land-MoE in comparison to state-of-the-art baseline methods.

\begin{figure}[htbp]
    \centering
	\includegraphics[ width=1.0\linewidth]{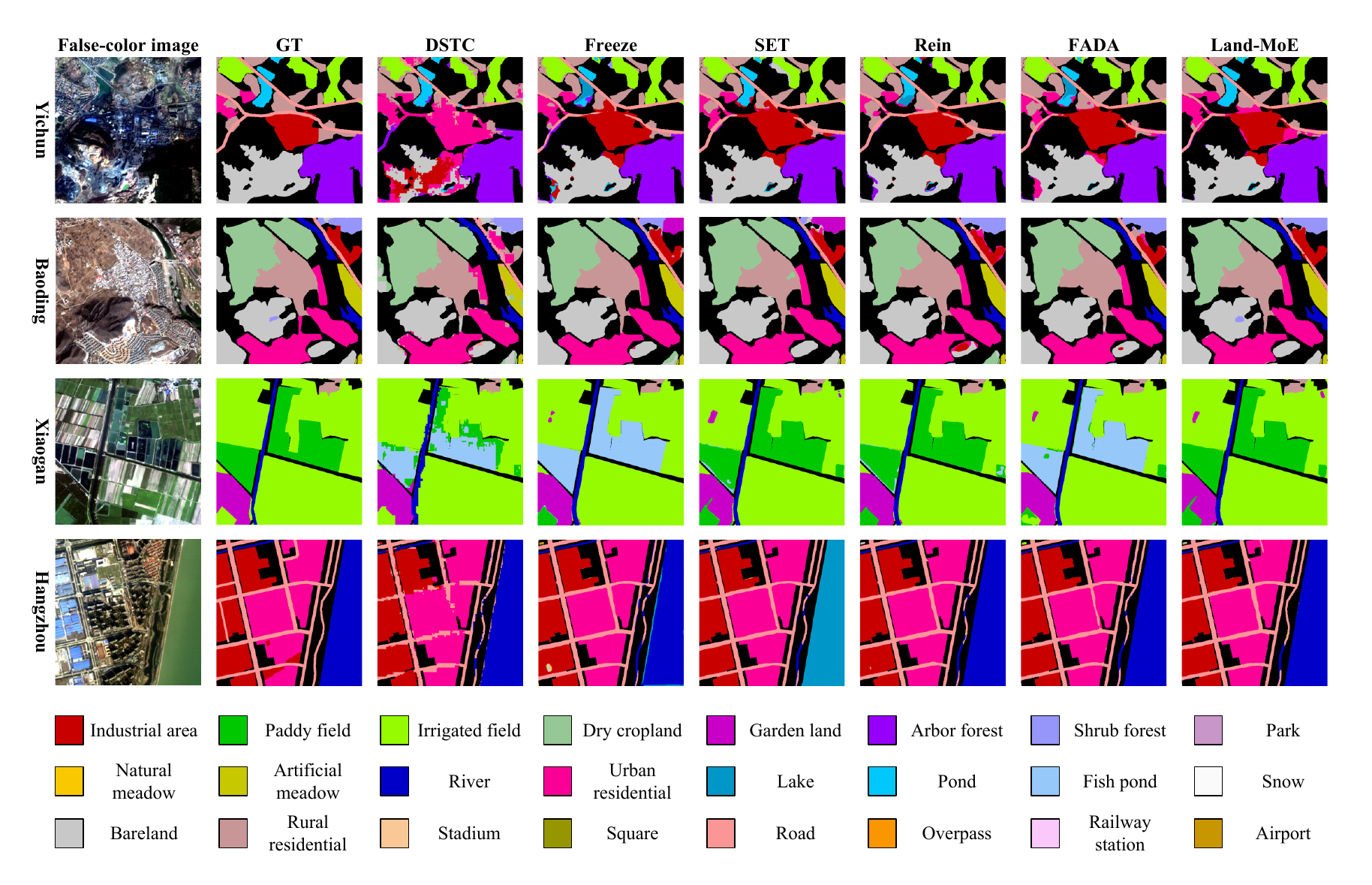}
        \caption{\textbf{Qualitative results showing predicted land cover classification maps for the cross-geospatial generalization task.} The figure illustrates the performance of Land-MoE in comparison to state-of-the-art baseline methods on cross-scene multispectral remote sensing images.}
    \label{fig:Cross_geospatial}
\end{figure}

Figure~\ref{fig:Cross_geospatial} further illustrates the cross-scene MLCC performance through predicted land cover maps for the aforementioned methods in the cross-geospatial generalization task.

\begin{figure}[htbp]
    \centering
	\includegraphics[ width=1.0\linewidth]{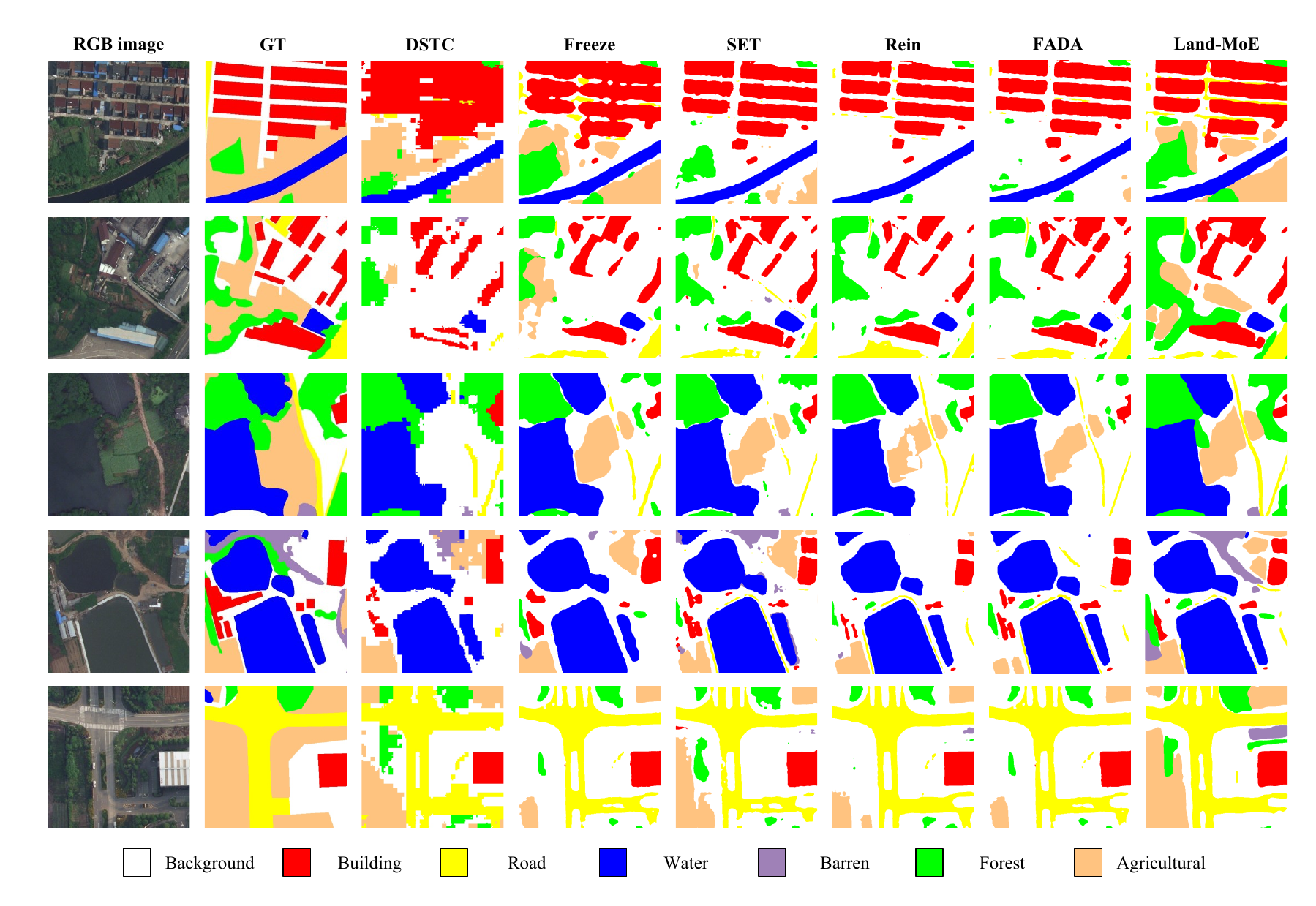}
        \caption{\textbf{Qualitative results showing predicted land cover classification maps for the Rural2Urban cross-scene task.} The figure compares the performance of Land-MoE with leading baseline methods on natural remote sensing images.}
    \label{fig:Rural2Urban}
\end{figure}

\begin{figure}[htbp]
    \centering
	\includegraphics[ width=1.0\linewidth]{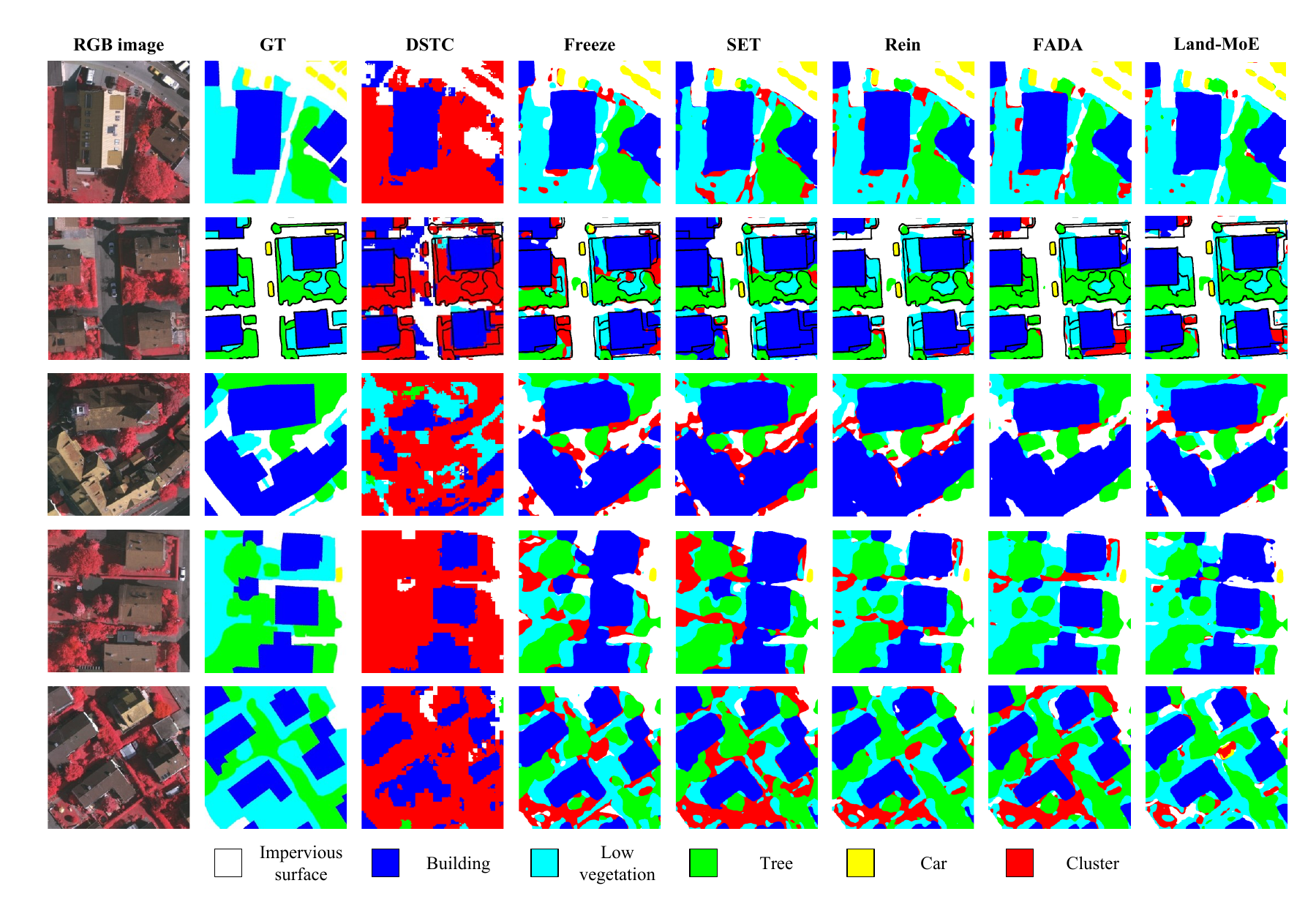}
        \caption{\textbf{Qualitative results showing predicted land cover classification maps for the Potsdam2Vaihingen cross-scene task.} The figure compares the performance of Land-MoE with leading baseline methods on natural remote sensing images.}
    \label{fig:Potsdam2Vaihingen}
\end{figure}

Figures~\ref{fig:Rural2Urban} and~\ref{fig:Potsdam2Vaihingen} respectively provide predicted cross-scene land cover classification results for Land-MoE and the leading baseline methods in the Rural2Urban and Potsdam2Vaihingen natural remote sensing image scenarios.

Collectively, these visual results demonstrate that Land-MoE consistently produces more accurate and coherent land cover classification maps compared to baseline methods across all presented cross-scene tasks, thereby qualitatively validating its excellent cross-scene generalization performance.

\end{document}